\documentclass[10pt,twocolumn,letterpaper]{article}

\usepackage[pagenumbers]{cvpr} 

\usepackage{booktabs}
\usepackage{animate}
\usepackage{enumitem}
\usepackage{multirow}
\usepackage{array}
\usepackage{bm}
\usepackage{makecell}
\usepackage[ruled]{algorithm2e}
\usepackage{marvosym}
\usepackage{url}
\usepackage[pagebackref=true,breaklinks=true,letterpaper=true,colorlinks,bookmarks=false]{hyperref}
\title{Generative Inbetweening through Frame-wise Conditions-Driven \\ Video Generation}

\author{Tianyi Zhu\textsuperscript{1} \ \ \ \ Dongwei Ren\textsuperscript{1, \Letter}\ \ \ \ Qilong Wang\textsuperscript{2} \ \ \ \  Xiaohe Wu\textsuperscript{1} \ \ \ \  Wangmeng Zuo\textsuperscript{1}\\
\textsuperscript{1}Harbin Institute of Technology  \ \ \ \
   \textsuperscript{2}Tianjin University
}

\begin{document}
\maketitle
\begin{abstract}
Generative inbetweening aims to generate intermediate frame sequences by utilizing two key frames as input. Although remarkable progress has been made in video generation models, generative inbetweening still faces challenges in maintaining temporal stability due to the ambiguous interpolation path between two key frames. This issue becomes particularly severe when there is a large motion gap between input frames. In this paper, we propose a straightforward yet highly effective Frame-wise Conditions-driven Video Generation (FCVG) method that significantly enhances the temporal stability of interpolated video frames. Specifically, our FCVG provides an explicit condition for each frame, making it much easier to identify the interpolation path between two input frames and thus ensuring temporally stable production of visually plausible video frames. To achieve this, we suggest extracting matched lines from two input frames that can then be easily interpolated frame by frame, serving as frame-wise conditions seamlessly integrated into existing video generation models. In extensive evaluations covering diverse scenarios such as natural landscapes, complex human poses, camera movements and animations, existing methods often exhibit incoherent transitions across frames. In contrast, our FCVG demonstrates the capability to generate temporally stable videos using both linear and non-linear interpolation curves. Our project page and code are available at \url{https://fcvg-inbetween.github.io/}.

\end{abstract}    
\begin{figure*}[!t]\footnotesize
        \centering
        \setlength{\tabcolsep}{1.5pt}
    \setlength{\abovecaptionskip}{0pt} 
    \setlength{\belowcaptionskip}{0pt}

    \hspace{-1em}
    \begin{tabular}[b]{cccc}
         \animategraphics[width=0.24\linewidth, autoplay, loop]{10}{imgs/mandance/film/}{0}{24}  & \animategraphics[width=0.24\linewidth, autoplay, loop]{10}{imgs/mandance/trf/}{0}{24} & 
         \animategraphics[width=0.24\linewidth, autoplay, loop]{10}{imgs/mandance/gi/}{0}{24} & \animategraphics[width=0.24\linewidth, autoplay, loop]{10}{imgs/mandance/ours/}{0}{24} \\
         \animategraphics[width=0.24\linewidth, autoplay, loop]{10}{imgs/scene/56_film/}{0}{24}  & \animategraphics[width=0.24\linewidth, autoplay, loop]{10}{imgs/scene/56_trf/}{0}{24} & 
         \animategraphics[width=0.24\linewidth, autoplay, loop]{10}{imgs/scene/56_gi/}{0}{24} & \animategraphics[width=0.24\linewidth, autoplay, loop]{10}{imgs/scene/56_ours/}{0}{24} \\
         \parbox[t][1.0em][t]{0.24\linewidth}{\centering FILM} & \parbox[t][1.0em][t]{0.24\linewidth}{\centering TRF} &
         \parbox[t][1.0em][t]{0.24\linewidth}{\centering GI} & \parbox[t][1.0em][t]{0.24\linewidth}{\centering Ours} \\

    \end{tabular}
    \caption{Video results of FILM \cite{reda2022film}, TRF \cite{feng2024explorative}, GI \cite{wang2024generative}, and our FCVG. These results are presented as animated videos that can be viewed in Adobe PDF reader. More videos are provided in the \href{https://fcvg-inbetween.github.io/}{project page}.}

    \label{fig:intro}
\end{figure*}

\section{Introduction}
\label{sec:intro}

Given the presence of low framerate videos, the generation of high framerate videos has emerged as an active research area with a wide range of applications that demands smooth and stable visual contents. Video interpolation or inbetweening, which focuses on synthesizing intermediate frames between two given frames, has been extensively studied in the literature \cite{li2023amt, Liu_2024_CVPR,danier2024ldmvfi, shen2024dreammover, zhou2023exploring}. However, traditional video interpolation methods are inherently limited in dealing with significant motions due to their reliance on optical flow for modeling frame motion. Recently, generative image-to-video (I2V) models have made remarkable progress in generating coherent videos \cite{blattmann2023stable,peng2024controlnext}, offering potential solutions to enhance framerates through generative inbetweening \cite{wang2024generative,yang2024vibidsampler}.

When start and end frames are provided, it is a straightforward task to separately generate two coherent videos using an I2V model, but the challenge of inbetweening lies in the ambiguity of the interpolation path caused by large motions, as shown in \cref{fig:path}(a). To address this issue, a time reversal sampling strategy \cite{feng2024explorative} is proposed to average the fusion of bidirectional diffusion denoising steps conditioned on start and end frames. Then, temporal attention layers in the I2V model are fine-tuned to enhance motion coherence \cite{wang2024generative}. More recently, Yang et al. \cite{yang2024vibidsampler} introduced a multi-channel sampling strategy to substitute direct average fusion. 
However, the ambiguity of interpolation path remains severe and often results in incoherent transitions in generated videos, as shown in \cref{fig:intro}.

In this paper, our aim is to mitigate the ambiguity in interpolation path and achieve temporal stability in video generation. 
We propose a straightforward yet highly effective Frame-wise Conditions-driven Video Generation (FCVG) method that significantly enhances the temporal stability of interpolated video frames. Specifically, our FCVG model provides an explicit condition when generating each frame using an I2V model, making it easier to identify the interpolation path between start and end frames, and thus ensuring temporally stable production of visually plausible video frames.
Firstly, we extract two conditions from the input key frames to establish robust correspondences between start and end frames using matched lines. In addition, pose skeletons can be incorporated into the conditions to better capture human poses. Subsequently, linear interpolation is employed on a frame-by-frame basis to interpolate the start and end conditions.
These frame-wise conditions effectively alleviate ambiguity in determining the interpolation path, and can be seamlessly integrated into the I2V model as control for video frame generation.

The linear frame-wise conditions for generative inbetweening are generally feasible from two perspectives:
(\emph{i}) In pioneering video interpolation methods \cite{jiang2018super,baker2011database,liu2017video}, the linear assumption is commonly adopted, which may not truly align with ground-truth temporal consistency but can lead to temporally stable videos for most scenes. As depicted in Fig. \ref{fig:intro}, our FCVG demonstrates a significantly higher level of video stability compared to existing methods;
(\emph{ii}) As illustrated in \cref{fig:path}(b), our frame-wise conditions provide a control path for inbetweening, while still allowing some flexibility for video generation models, whose influence can be further adjusted by tuning the fusion weight between features of condition and video generation branches. 
Moreover, our FCVG allows users to specify a non-linear interpolation path for frame-wise conditions to generate desired video frames, providing more flexibility in determining the interpolation path.

Extensive experiments are conducted on the collected diverse testing samples including natural landscapes, complex human poses, camera movements and animations, where our FCVG is compared with both video interpolation methods and generative inbetweening methods. 
Regarding several evaluation metrics, our method outperforms existing methods in terms of frame textures and temporal stability. 
Particularly when dealing with large motions, unlike other methods that suffer from incoherent transitions across frames, our FCVG exhibits significantly enhanced temporal stability in the generated videos.

\section{Related Work}

\noindent {\bf Optical Flow-based Frame Interpolation.}
Video frame interpolation aims to synthesize intermediate frames between two given input frames \cite{bao2019depth,niklaus2017video}. 
Previous methods primarily rely on optical flow-based approaches\cite{li2023amt, Hu_2024_CVPR, Liu_2024_CVPR}, which estimate optical flow and apply forward \cite{niklaus2020softmax} or backward warping to generate intermediate frames. 
Some techniques incorporate flow reversal techniques to estimate the intermediate flows \cite{wu2024perception, sim2021xvfi, jiang2018super}, while others concentrate on directly predicting the intermediate flows \cite{huang2022real, kong2022ifrnet, zhang2023extracting, zhong2024clearer}. 
Although these methods yield stable interpolation results in real-world scenes, they exhibit significant artifacts when confronted with large motion or complex scenarios, such as human motion. 
Moreover, they struggle with the impact of optical flow estimation accuracy \cite{chen2022improving} in varying data distributions, such as animation or line art.
While some approaches have developed models tailored for animation \cite{chen2022improving, siyao2021deep} or line art \cite{siyao2023deep, shen2024bridging, zhu2024thin}, restricting them for specific data types.

\noindent {\bf Diffusion-based Frame Interpolation.}
In recent years, diffusion models have demonstrated remarkable capabilities in generating high-quality images and videos \cite{rombach2022high, ho2020denoising, blattmann2023align}. Several studies have explored the effectiveness of diffusion models for video frame interpolation \cite{danier2024ldmvfi, shen2024dreammover, voleti2022mcvd, huang2024motion}, particularly in addressing complex motions that pose significant challenges for optical flow-based methods. 
Some approaches treat the input frames as conditions and utilize large-scale data to train a diffusion model for video frame interpolation \cite{jain2024video, xing2025dynamicrafter,xing2024tooncrafter}. Other approaches leverage pre-trained image-to-video diffusion models, incorporating new sampling strategies to achieve frame interpolation \cite{feng2024explorative, wang2024generative, yang2024vibidsampler}. TRF \cite{feng2024explorative} proposes a time reversal sampling strategy that fuses bidirectional motion from two parallel diffusion denoise steps conditioned on the start and end frame. Based on the time reversal strategy, Generative Inbetweening \cite{wang2024generative} fine-tunes diffusion models by utilizing temporal attention information to maintain motion consistency, while VIBIDSampler \cite{yang2024vibidsampler} introduces a bidirectional sampling approach rather than direct average fusion ensuring on-manifold generation of intermediate frames.

\noindent {\bf Controllable Video Generation.}
Recent works try to introduce controllable conditions to video generation models \cite{zhang2024controlvideo}, such as camera motion \cite{he2024cameractrl, wang2024motionctrl} and human pose \cite{zhang2024mimicmotion,zhu2024champ, hu2024animate}. Among these methods, the approach utilizing lightweight adapters \cite{mou2024t2i, zhang2023adding} is preferred by many researchers due to its elimination of the need for pre-training the large diffusion model. 
ControlNet \cite{zhang2023adding} uses zero convolution to connect certain trainable layers copied from pre-trained large models to the original layers. To reduce the additional computational cost, ControlNeXt \cite{peng2024controlnext} introduces a lightweight module and fine-tunes several parameters in the diffusion model, aligning them using cross normalization.

\section{Method}

\begin{figure}[tb]
  \centering
 \setlength{\abovecaptionskip}{0pt} 
  \setlength{\belowcaptionskip}{0pt}
  \includegraphics[width=\linewidth]{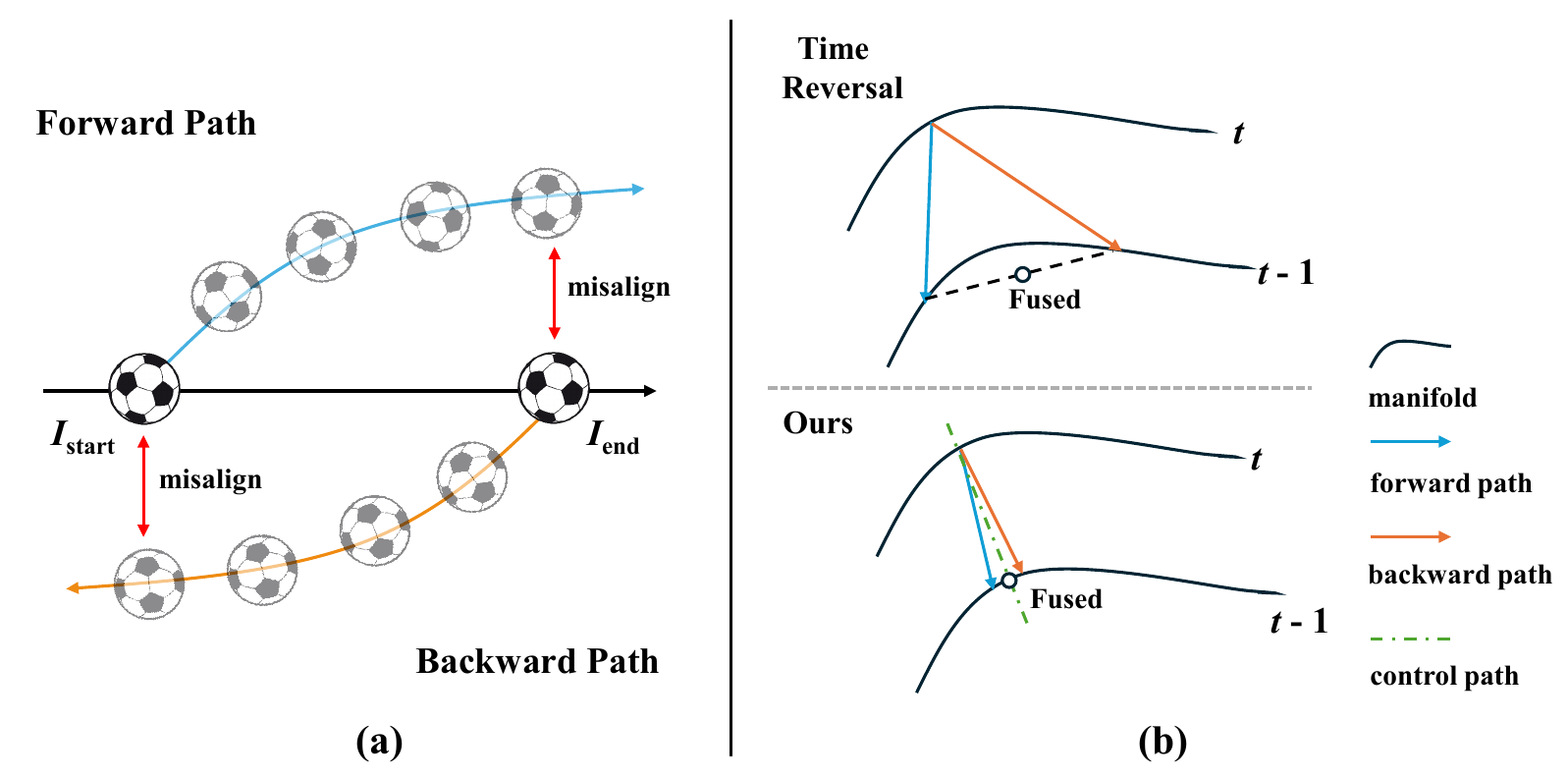}
  \caption{(a) The ambiguity in interpolation path, where the stochastic nature of forward and backward paths leads to misalignment or even generating substantially different contents in two paths. (b) Our frame-wise conditions serve as a control path, enabling rough alignment of the forward and backward paths, thereby confining the fusion process closer to the manifold.}
  \label{fig:path}
\end{figure}

\subsection{Preliminaries}

Diffusion model \cite{ho2020denoising} is a type of generative model that uses a denoising network to iteratively denoise random Gaussian noise to generate a high-quality image or video. Specifically, for I2V diffusion models, such as Stable Video Diffusion (SVD) \cite{blattmann2023stable}, given a video $\bm x \in \mathbb{R}^{N \times 3 \times H \times W}$ containing $N$ frames, SVD first encode $\bm x$ using an autoencoder $\mathcal{E}(\cdot)$ to get latent representation $\bm z \in \mathbb{R}^{N \times C \times {H} \times {W}}$ \cite{vaeKingmaW13}. The forward process of diffuse gradually adds noise to latent representation $\bm z$ as follows
\begin{equation}
    \bm z_t = \alpha_t \bm z + \sigma_t \bm \epsilon, 
\end{equation}
where $\bm \epsilon \sim \mathcal{N}(\bm 0, \bm I)$, $\alpha_t$ and $\sigma_t$  represent the noise level for denoising time $t$, defined by the noise schedule. As for backward process, a 3D UNet \cite{ronneberger2015u} denoiser $f_{\theta}$ is used for iteratively denoising under the image condition $\bm c_{\text{image}}$. The optimization objective is formulated as
\begin{equation}
    \mathcal{L} = \mathbb{E}_{\bm z, \bm c_{\text{image}}, \bm \epsilon \sim \mathcal{N}(\bm 0, \bm I), t} \left[ \left \| \bm v - f_{\theta}(\bm z_t, \bm c_{\text{image}}, t)\right \|^{2}_{2} \right],
\end{equation}
where $\bm v = \alpha_t \bm \epsilon_t - \sigma_t \bm z_t$ is referred as v-prediction \cite{salimans2022progressive}.
Finally, the generated videos are obtained through the VAE decoder $\hat{\bm x} =\mathcal{D}(\bm z_0)$.

Inspired by the strong capability of diffusion models in video generation, several studies have endeavored to integrate diffusion models into video inbetweening to tackle challenges posed by intricate and extensive motions that are difficult to address using optical flow-based methods \cite{danier2024ldmvfi, huang2024motion, feng2024explorative}. A direct approach for applying diffusion models in video inbetweening involves utilizing key frames as conditions for retraining the diffusion model. However, this often necessitates substantial data and computational resources and may also be influenced by discrepancies in data distribution. Consequently, some approaches leverage pre-trained diffusion models for video inbetweening \cite{feng2024explorative, wang2024generative, yang2024vibidsampler}. The fundamental concept behind these methods is to fuse the temporal forward path and backward path after each denoising step conditioned on start and end frames, respectively, i.e., a process referred to ``time reversal" \cite{feng2024explorative, wang2024generative}.

\begin{figure}[tb]
  \centering
 \setlength{\abovecaptionskip}{0pt} 
  \setlength{\belowcaptionskip}{0pt}
  \includegraphics[width=\linewidth]{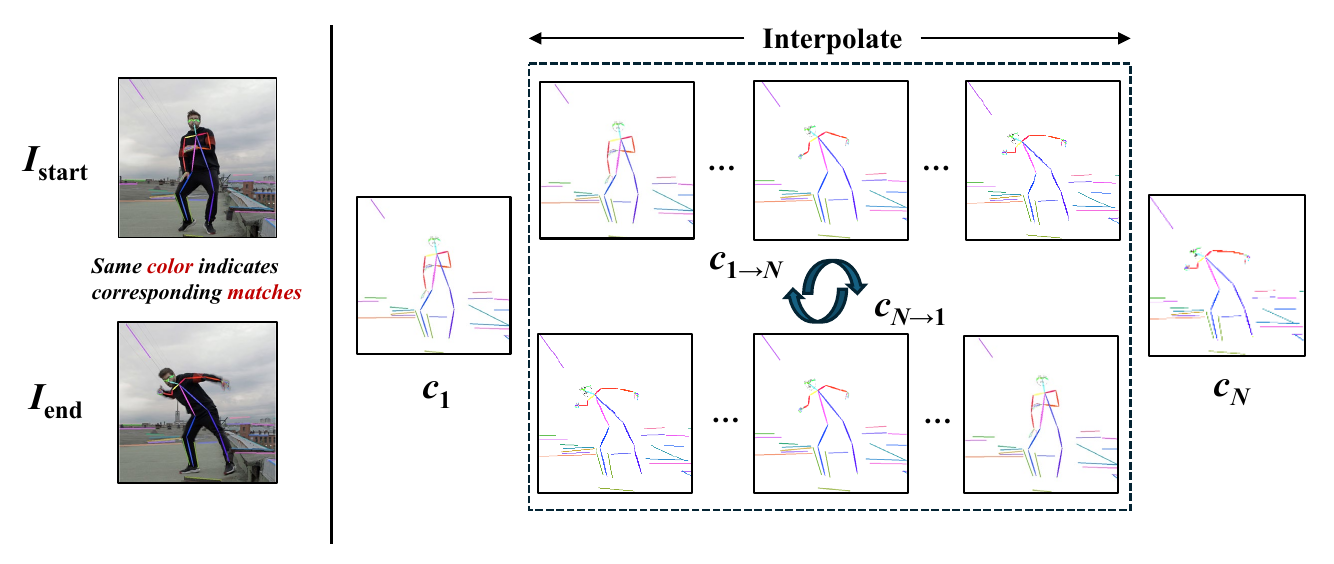}
  \caption{The process for acquiring forward and backward frame-wise conditions $\bm{c}_{1\rightarrow N}$ and $ \bm{c}_{N\rightarrow 1}$. Two initial conditions $\bm{c}_1$ and $\bm{c}_N$ can be obtained by establishing correspondence between start frame $\bm{I}_\text{start}$ and end frame $\bm{I}_\text{end}$, where the same color indicates corresponding matches. 
  Then, frame-wise conditions are obtained by interpolating $\bm{c}_1$ and $\bm{c}_N$.}
  \label{fig:condition_interpolate}
\end{figure}

\subsection{Motivation of Frame-wise Conditions} \label{sec:framecondition}

Although time reversal provides a means to directly utilize pre-trained video generations inbetweening, it exhibits certain limitations \cite{feng2024explorative} that are summarized as follows: (\emph{\textbf{i}}) The motion generated by I2V models tends to be diverse rather than stable. While this diversity is advantageous for pure I2V tasks, it introduces significant ambiguity when applying the time reversal strategy for video inbetweening. As depicted in \cref{fig:path}(a), the stochastic nature of generating forward and backward paths leads to misalignment and even substantially different contents in two paths, resulting in an unstable and unrealistic videos. (\emph{\textbf{ii}}) Tedious tuning of hyper-parameters related to temporal conditioning within the I2V model is required for each input pair, such as motion bucket ID and frames per second.  
(\emph{\textbf{iii}}) Inference efficiency is constrained by certain techniques, e.g., noise re-injection \cite{feng2024explorative}, aimed at mitigating the ambiguity but significantly increasing inference time (approximately 1.5 to 3 times longer).

Indeed, if the first fundamental limitation in ambiguity can be addressed, the subsequent two issues can also be readily resolved. To this end, several studies have made significant efforts in mitigating the misalignment between forward and backward paths \cite{wang2024generative,yang2024vibidsampler}. 
Nevertheless, as depicted in \cref{fig:path}(b), there still exists considerable stochasticity between these paths, thereby constraining the effectiveness of these methods in handling scenarios involving large motions such as rapid changes in human poses.
The ambiguity in the interpolation path primarily arises from insufficient conditions for intermediate frames, since two input images only provide conditions for start and end frames.
Therefore, in this work, we suggest offering an explicit condition for each frame, which significantly alleviates the ambiguity of the interpolation path. 
As shown in \cref{fig:path}(b), frame-wise conditions ensure that the forward and backward paths are relatively aligned during the denoising process, thereby rendering a simple fusion method adequate to confine the fusion process closer to the manifold.

\begin{figure}[tb]
  \centering
    \setlength{\abovecaptionskip}{0pt} 
  \setlength{\belowcaptionskip}{0pt}
  \includegraphics[width=.95\linewidth]{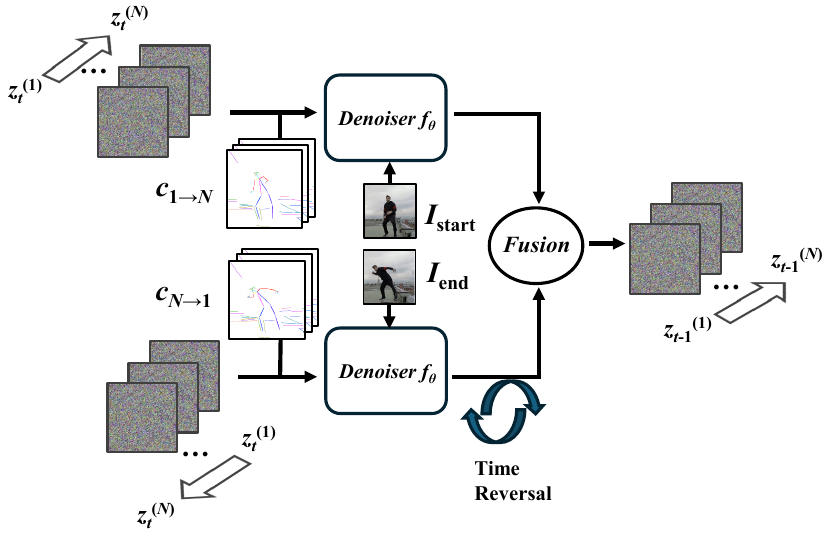}
  \caption{Inference of FCVG at time $t$.}
  \label{fig:model_timereverse}
\end{figure}

\subsection{Proposed FCVG}

Given two input frames $\bm I_{\text{start}}$ and $\bm I_{\text{end}}$, our FCVG aims to generate $N$ video frames, whose start and end frames should be consistent with $\bm I_{\text{start}}$ and $\bm I_{\text{end}}$, respectively. 
As shown in \cref{fig:model_timereverse}, our FCVG provides frame-wise conditions for video generation model to make the generated frames be temporally stable. 
Based on the time reversal strategy, our FCVG at time $t$ can be specifically formulated as
\begin{equation}
     \Tilde{\bm{z}}_t = f_{\theta}\left(\bm z_{t+1}, \bm I_{\text{start}}, \bm c_{1\rightarrow N}, t\right),
\end{equation}
\begin{equation}
    \bm \Tilde{z}'_t = f_{\theta}\left(\text{flip}(\bm z_{t+1}), \bm I_{\text{end}},\bm c_{N \rightarrow 1}, t\right),
\end{equation}
\begin{equation}
    \bm z_t = \bm{\lambda} \cdot \bm \Tilde{z}_t+ (1- \bm{\lambda} ) \cdot \text{flip}(\bm \Tilde{z}'_t),
\end{equation}
where $\bm z_{t+1}\in \mathbb{R}^{N\times C\times H \times W}$ is diffusion noises from time $t+1$, 
$f_{\theta}$ is the pre-trained denoiser model with control module,
 $\bm c_{1 \rightarrow N}, \bm c_{N \rightarrow 1} \in \mathbb{R}^{N\times 3\times H \times W}$ are temporal aligned forward and backward frame-wise conditions, $\text{flip}(\cdot)$ denotes flipping the sample along the time dimension, and $\bm \lambda \in \mathbb{R}^{N}$ is fusion weights with $\lambda_i = 1 -\frac{i-1}{N-1}, i \in \{1,...,N\}$. 
 The symbol $\cdot$ denotes frame-wise multiplication, i.e., $\lambda_i$ is multiplied with noises for $i$-th frame. 
By iteratively denoising until time $t=0$, final inbetweening video frames can be obtained by $\hat{\bm x}_0 = \mathcal{D}(\bm z_0)$.

\begin{algorithm}[tb]
\small

\SetKwComment{Comment}{/* }{ */}
\SetKwInput{KwData}{Input}
\SetKwInput{KwResult}{Return}
\caption{Inference of FCVG}\label{alg:sample}
\KwData{$\bm I_{\text{start}}, \bm I_{\text{end}}, f_{\theta}, \mathcal{D}, \bm z_T \sim \mathcal{N}(\mathbf{0}, \mathbf{I})$}

Computing $\bm{\lambda}$ with $\lambda_i = 1 -\frac{i-1}{N-1}, i \in \{1,...,N\}$\;
Extracting conditions $\bm{c}_{1}, \bm{c}_{N}$ from 
$\bm{I}_{\text{start}}, \bm{I}_{\text{end}}$\;

$\bm c_{1 \rightarrow N} = \text{interpolate}(c_1, c_N)$\;
$\bm c_{N \rightarrow 1} = \text{flip}(\bm c_{1 \rightarrow N})$\;
 \For{$t\gets T$ $:$ $1$}{
     $\bm \Tilde{\bm z}_t = f_{\theta}\left(\bm z_{t}, \bm I_{\text{start}}, \bm c_{1\rightarrow N}, t\right)$\; 

     $\bm z_{t, f} = \text{flip}(\bm z_t)$\;

     $\bm \Tilde{\bm z}_t' = f_{\theta}\left(\bm z_{t, f}, \bm I_{\text{end}}, \bm c_{N\rightarrow 1}, t\right)$\; 
     $\Tilde{\bm z}_{t, f}' = \text{flip}(\Tilde{\bm z}_t')$\;
     $\bm z_{t-1} = \bm \lambda \cdot \bm \Tilde{z}_t+ (1-\bm \lambda)\cdot \Tilde{\bm z}_{t, f}'$;
}
\KwResult{$\mathcal{D}(\bm z_0$)}
\end{algorithm}

\subsubsection{Frame-wise Conditions}

Currently, we only have access to two image conditions, i.e., $\bm{I}_\text{start}$ and $\bm{I}_\text{end}$. However, extending them as frame-wise conditions is infeasible.  
In order to acquire frame-wise conditions, the initial conditions should satisfy two properties that can effectively capture frame motion, and are amenable for extension as frame-wise.
Drawing inspiration from prior works \cite{Liu_2024_CVPR, zhu2024thin} that showcase the robustness of global matching in handling large motions and complex scenes, we propose to employ the line matching model to extract the initial conditions from $\bm{I}_\text{start}$ and $\bm{I}_\text{end}$. Specifically, we utilize the pre-trained GlueStick \cite{pautrat2023gluestick} as our line matching model to establish correspondences between $\bm{I}_\text{start}$ and $\bm{I}_\text{end}$, and subsequently visualize these matches as images with distinct colors representing different line matches, resulting in two initial conditions $\bm c_{1}$ and $\bm c_{N}$.
Moreover, to further improve human motions, we extract human poses that can be directly added into $\bm c_{1}$ and $\bm c_{N}$. 

As shown in \cref{fig:condition_interpolate}, the control condition $\bm c_{i}$ for $i$-th frame can be obtained by accordingly interpolating $\bm c_{1}$ and $\bm c_{N}$, where the forward frame-wise conditions $\bm c_{1 \rightarrow N}$ are the concatenation of $\{\bm c_{i}\}_{i=1}^N$ along the time dimension, and backward frame-wise conditions $\bm c_{N \rightarrow 1}$ are the flip of $\bm c_{1 \rightarrow N}$ along the time dimension.
We empirically found that linear interpolation is sufficient for most cases to guarantee temporal stability in inbetweening videos, and our method allows users to specify non-linear interpolation paths for generating desired videos, referring to \cref{fig:nonlinear}.

\subsubsection{Injection to Video Generation Model}

\begin{figure}[tb]
  \centering
  \includegraphics[width=0.85\linewidth]{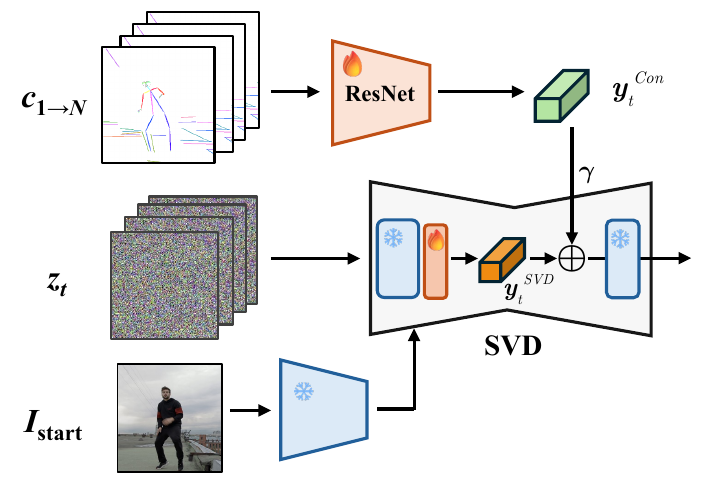}
  \caption{Overview of injecting frame-wise conditions into SVD. To make frame-wise conditions better fit pre-trained SVD, we only need to fine-tune a small set of parameters including the value and output projection matrices within the attention layers, and the lightweight ResNet blocks.}
  \label{fig:model_finetune}
  \vspace{-3mm}
\end{figure}

We follow ControlNeXt \cite{peng2024controlnext} to inject frame-wise conditions into the I2V model. We choose SVD \cite{blattmann2023stable} as our base I2V model. Compared to other controllable video generation methods, ControlNeXt is light-weight and does not significantly increases inference time. Specifically, the control conditions are first encoded by a lightweight module composed of multiple ResNet \cite{he2016deep} blocks. These encoded conditions are first processed using cross normalization to align the distributions of $\bm y^\text{Con}$ and $\bm{y}^{\text{SVD}}$ from the condition and SVD branches. Then at denoising time $t$, frame-wise conditions can be injected into SVD as follows

\begin{equation}
   \hat{\bm{y}}_t^{} = {\bm{y}}_t^{\text{SVD}} + \gamma \  {\bm{y}}_t^\text{Con},
\end{equation}
where $\gamma$ is a tunable weight for controlling the significance of $\bm y^\text{Con}$. 
The inference of FCVG is detailed in \cref{alg:sample}.
Moreover, in order to enhance the compatibility of pre-trained SVD with our frame-wise conditions, we employ a small set of videos for fine-tuning. We freeze the majority of SVD parameters and solely optimize the value and output projection matrices within attention layers, along with lightweight ResNet blocks, as shown in \cref{fig:model_finetune}.

The limitations in \cref{sec:framecondition} have been largely resolved in FCVG:
(\emph{\textbf{i}}) By explicitly specifying the condition for each frame, the ambiguity between forward and backward paths is significantly alleviated;
(\emph{\textbf{ii}}) Only one tunable parameter $\gamma$ is introduced, and setting it to 1, while keeping hyper-parameters in SVD as default, yields favorable results in most scenarios; 
(\emph{\textbf{iii}}) A simple average fusion, without noise re-injection, is adequate in FCVG, and the inference steps can be substantially reduced by 50\% compared to GI \cite{wang2024generative}.

\section{Experiments}

\begin{table*}[tb]
 \setlength{\abovecaptionskip}{3pt} 
  \setlength{\belowcaptionskip}{0pt}
  \setlength{\tabcolsep}{1.3mm}
  \caption{Quantitative comparison on different interpolation gaps.
  Our primary focus lies in generative approaches, with FILM being the sole method that employs optical flow.
  \textbf{Bold} refer to the best results.
  All these metrics are not capable of precisely evaluating temporal stability of generated videos, and thus we highly recommend directly observing video results.
  }
  \label{tab:compare}
  \centering
   \small
  \begin{tabular}{@{}lccccc|ccccc@{}}
    \toprule
     Method & \multicolumn{5}{c}{$\text{Frame Gap}=23$} & \multicolumn{5}{c}{$\text{Frame Gap}=12$}\\
     & LPIPS ($\downarrow$) & FID ($\downarrow$)  & VBench ($\uparrow$) & FVMD ($\downarrow$) & FVD ($\downarrow$) & LPIPS ($\downarrow$)& FID ($\downarrow$)   & VBench ($\uparrow$) & FVMD ($\downarrow$) & FVD ($\downarrow$) \\

     \specialrule{.05em}{.4ex}{.65ex}
     FILM\cite{reda2022film} & \textbf{0.1540} & 25.00  & 0.8615 & 8208.7 & 543.4 & \textbf{0.1980}& 24.44  & 0.8667 & 6975.9 & 495.4\\

    \specialrule{.05em}{.4ex}{.65ex}
    DynamiCrafter\cite{xing2025dynamicrafter} & 0.3886 & 52.66  & 0.8410 & 13221.9 & 978.9 & 0.3839& 37.49  & 0.8458 & 11810.7& 652.5\\
    TRF\cite{feng2024explorative} & 0.3687 & 42.76  &0.8438 & 10458.0&823.4 & 0.3742& 39.01  & 0.8478 & 10076.6 & 818.4\\
    GI\cite{wang2024generative} & 0.2155 & 31.39& 0.8606& 5682.6&524.0& 0.2615 & 32.37  & 0.8651 & 4721.0 & 565.8\\
    \specialrule{.05em}{.4ex}{.65ex}
    Ours & 0.1832 & \textbf{24.05}& \textbf{0.8619}& \textbf{5607.2}& \textbf{437.9} & 0.2378 & \textbf{22.77}  & \textbf{0.8672} & \textbf{4537.4} & \textbf{465.6}\\
  \bottomrule
  \end{tabular}
\end{table*}

\begin{figure*}[tb]
  \centering
    \setlength{\abovecaptionskip}{0pt} 
  \setlength{\belowcaptionskip}{3pt}
  \includegraphics[width=\linewidth]{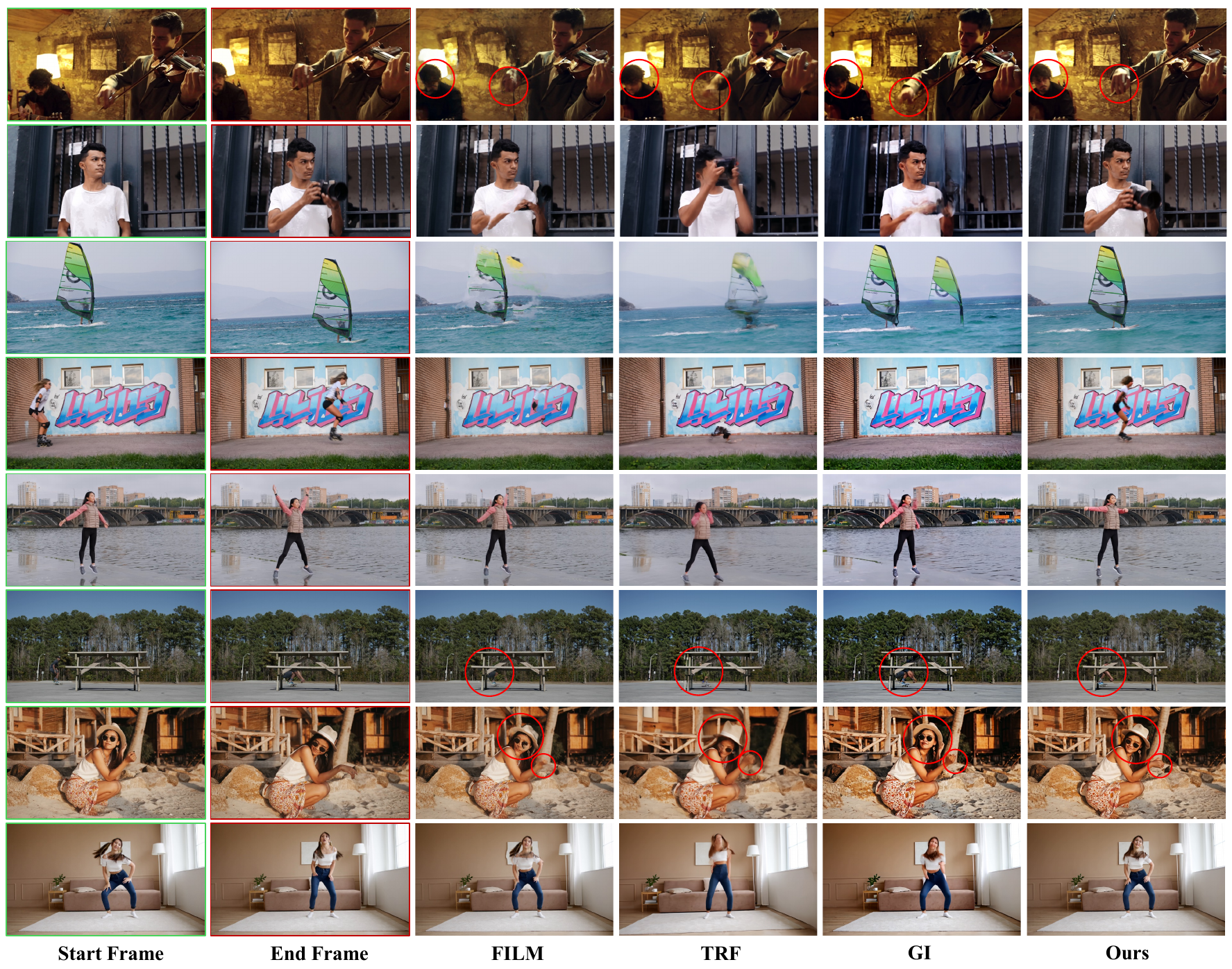}
  \caption{Qualitative evaluation on diverse scenes, where our FCVG is superior in texture details and coherent intermediate motions.}
  \label{fig:results compare}
  \vspace{-3mm}
\end{figure*}

\subsection{Experimental Setup}
\noindent{\textbf{Datasets.}} 
To verify the performance of our FCVG across diverse scenes,
we collect a dataset encompassing a variety of scenes, such as natural environments, indoor/outdor scenes, and human poses, where diverse motion types are encompassed such as camera movements, object motion, human dance actions, and facial expression transitions. Specifically, our dataset consists of 524 video clips, each containing 25 frames, selected from the DAVIS dataset \cite{pont20172017} and RealEstate10K dataset \cite{zhou2018stereo}, supplemented by high-frame-rate videos from Pexels\footnote{https://www.pexels.com/}. We randomly split the dataset in a 4:1 ratio for fine-tuning and testing respectively.

\noindent{\textbf{Evaluation Metrics.}}
Following previous works \cite{wang2024generative, feng2024explorative}, we adopt LPIPS \cite{zhang2018unreasonable} and Fréchet Inception Distance (FID) \cite{heusel2017gans, parmar2022aliased} to evaluate the quality of individual frames, while employing Fréchet Video Distance (FVD) \cite{unterthiner2019fvd} to assess the overall quality of videos. 
Additionally, we take two recently proposed metrics VBench \cite{huang2024vbench} and FVMD \cite{liu2024fr} to assist the evaluation, where VBench assesses videos across multiple dimensions based on pre-trained models, while FVMD refines FVD by emphasizing more on motion consistency. 
Furthermore, it should be noted that all these metrics are not capable of precisely evaluating temporal stability of generated videos, and thus we highly recommend directly observing more video results provided in supplementary file.

\noindent{\textbf{Implementation Details.}}
For obtaining initial conditions, we utilize the pre-trained GlueStick \cite{pautrat2023gluestick} for line matching and DWPose \cite{yang2023effective} for estimating human poses. 
The fine-tuning is performed on ResNet blocks, and the value and output projection matrices within the attention layer of SVD. The fine-tuning process is conducted for 70k iterations using the AdamW optimizer on an NVIDIA A800 GPU, with a learning rate of $1\times 10^{-6}$ and $\beta_1 = 0.9, \beta_2 = 0.999$. For fine-tuning, we crop the frames to patches with resolution $512 \times 320$. 
The inference of FCVG is done in $T=25$ steps without noise re-injection. 
Without specific clarity, the balancing weight $\gamma=1$ is set for all the experiments. 
Due to the robustness of our FCVG, all the hyper-parameters in SVD adopt the default settings.

\begin{figure*}[tb]
  \centering
    \setlength{\abovecaptionskip}{0pt} 
  \setlength{\belowcaptionskip}{0pt}
  \includegraphics[width=\linewidth]{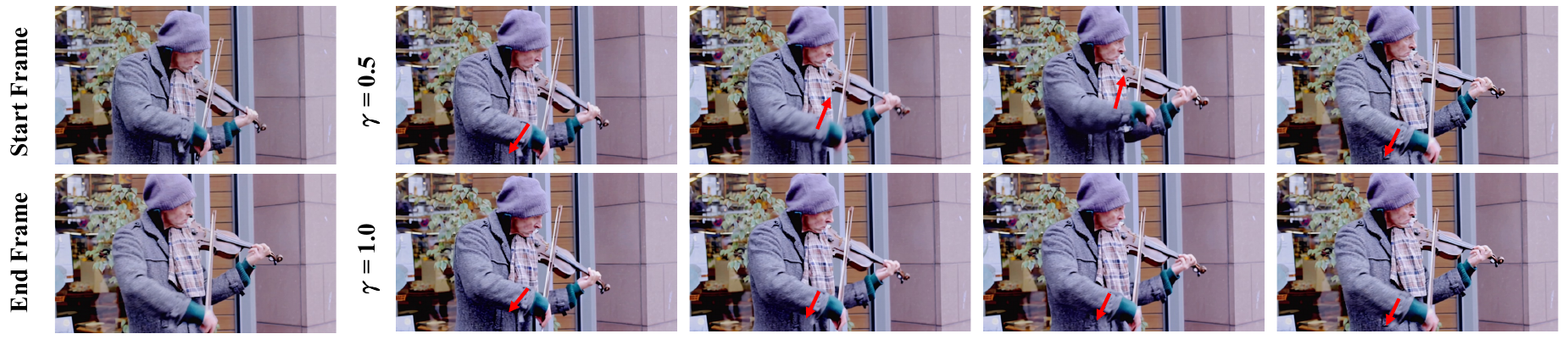}
  \caption{The effect of control weight $\gamma$. Red arrows indicate the movement directions. As the value of $\gamma$ decreases, the diversity of intermediate motion increases, e.g., up-and-down swinging of the arm, while temporal stability is guaranteed for two cases. The videos are available in the project page.}
  \label{fig:weight_compare}
  \vspace{-2mm}
\end{figure*}

\subsection{Comparison with State-of-the-arts}

We compare FCVG with state-of-the-art optical flow-based interpolation method  FILM \cite{reda2022film}, and diffusion-based methods including DynamiCrafter \cite{xing2025dynamicrafter}, TRF \cite{feng2024explorative} and GI \cite{wang2024generative}.

\vspace{1mm}\noindent {\bf Quantitative evaluation.}
To evaluate performance under different motion conditions, we conduct assessments with frame gaps setting to 23 and 12, respectively. 
As shown in \cref{tab:compare}, our method achieves the best performance among four generative approaches across all the metrics. 
Regarding the LPIPS comparison with FILM, our FCVG is marginally inferior, while demonstrating superior performance in other metrics. 
Considering the absence of temporal information in LPIPS, it may be more appropriate to prioritize other metrics and visual observation. 
Moreover, by comparing the results under different frame gaps, FILM may work well when the gap is small, while generative methods are more suitable for large gap.
Among these generative methods, our FCVG exhibits significant superiority owing to its explicit frame-wise conditions.

\noindent {\bf Qualitative evaluation.}
The visual comparison in \cref{fig:results compare} illustrates the superior performance of our method compared to other counterparts in terms of motion stability, consistency, and overall quality. While FILM produces smooth interpolation results for small motion scenarios, it struggles with large scale motion due to inherent limitations of optical flow, resulting in noticeable artifacts such as background and hand movement (in the first case). Generative models like TRF and GI suffer from ambiguities in fusion paths leading to unstable intermediate motion, particularly evident in complex scenes involving human and object motion. In contrast, our method consistently delivers satisfactory results across various scenarios. Even when significant occlusion is present (in the second case and sixth case), our method can still capture reasonable motion. Furthermore, our approach exhibits robustness for complex human actions (in the last case).

\noindent {\bf Computational efficiency.}
In \cref{tab:infer time}, we evaluate the inference time of diffusion-based methods. DynamiCrafter generates 16 frames at resolution of $512\times 320$, while the other methods generate 25 frames at resolution of $1024 \times 576$. DynamiCrafter exhibits advantages in inference time due to its single-pass denoising process. In comparison to TRF and GI, our FCVG facilitates easier alignment between forward and backward paths, ensuring that the results remain within the manifold as in \cref{sec:framecondition}. Consequently, although FCVG requires slightly less extra time for conditions extraction, it eliminates noise re-injection step and achieves satisfactory results with only 25 steps, in contrast to the requirement of 50 steps by the other methods. Real-time applications still pose challenges for generative inbetweening due to the high inference cost of pre-trained video generation models.

\begin{table}[htb]\small
 \setlength{\abovecaptionskip}{3pt} 
  \setlength{\belowcaptionskip}{0pt}
  \setlength{\tabcolsep}{1.5mm}
  \caption{{Evaluation of computational efficiency.}
  }
  \label{tab:infer time}
  \centering
  \begin{tabular}{@{}lccccc@{}}
    \toprule
     Methods & $N\times (H,W)$ & $T$  &  Time (\emph{s})\\
    \midrule
    DynamiCrafter \cite{xing2025dynamicrafter}& 16$\times (512,320)$ & 50 & 37 \\
    TRF \cite{feng2024explorative}& 25$\times (1024,576)$ & 50 & 1230\\
    GI \cite{wang2024generative} & 25$\times (1024,576)$ & 50 & 975 \\
    Ours & 25$\times (1024,576)$& 25 & 523 \\
  \bottomrule
  \end{tabular}
\end{table}

\subsection{Flexibility of Frame-wise Conditions}\label{sec:flexibility}
\begin{figure}[htb]
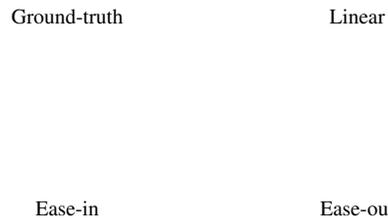
\footnotesize
        \centering
        \setlength{\tabcolsep}{1.5pt}
    \setlength{\abovecaptionskip}{0pt} 
    \setlength{\belowcaptionskip}{0pt}
    \begin{tabular}{cc}
         \animategraphics[width=0.45\linewidth, autoplay, loop]{10}{imgs/dif_motion/096_gt/}{0}{24}  & \animategraphics[width=0.45\linewidth, autoplay, loop]{10}{imgs/dif_motion/ours_curve/}{0}{24} \\ 
Ground-truth & Linear\\
\animategraphics[width=0.45\linewidth, autoplay, loop]{10}{imgs/dif_motion/easein_curve/}{0}{24} & \animategraphics[width=0.45\linewidth, autoplay, loop]{10}{imgs/dif_motion/easeout_curve/}{0}{24}\\
Ease-in & Ease-out \\
    \end{tabular}
    \caption{FCVG is able to handle linear and non-linear interpolation curves. The videos can be viewed in Adobe PDF reader.}
    \label{fig:nonlinear}
    \vspace{-3mm}
\end{figure}

Our FCVG allows some flexibility to SVD by setting different values for $\gamma$. In \cref{fig:weight_compare}, one can see that adjusting $\gamma$ within a certain range has little impact on video stability but can lead to different directions of movement.
Albeit fine-tuned for linear frame-wise conditions, FCVG enables users to specify non-linear motion trajectories. As depicted in \cref{fig:nonlinear}, FCVG can readily generates stable videos with ease-in and ease-out non-linear motion trajectories.
More video examples can be found in supplementary file.

\begin{figure*}[!t]
  \centering
    \setlength{\abovecaptionskip}{0pt} 
  \setlength{\belowcaptionskip}{3pt}
  \includegraphics[width=0.99\linewidth]{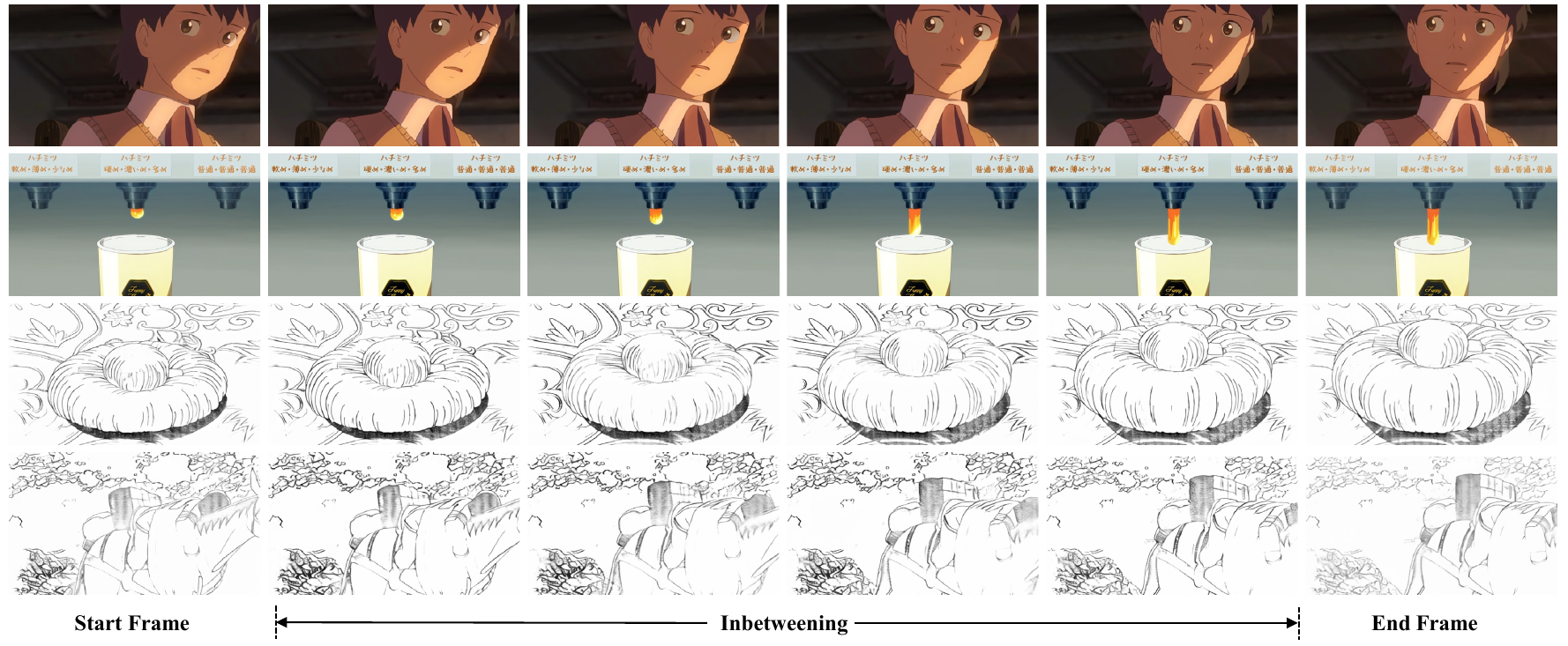}
  \caption{Inbetweening results on animations and linearts. FCVG exibits favorable performance without fine-tuning on these data types.}
  \label{fig:anime_results}
  \vspace{-3mm}
\end{figure*}

\subsection{Generalization to Animation Videos}
We further evaluate the generalization ability of FCVG to hangle animation and lineart videos, whose data types do not appear in the fine-tuning dataset. As depicted in \cref{fig:anime_results}, our FCVG consistently produces visually appealing results, even for challenging scenarios such as head turns, liquid flow, and object deformations. This can be attributed to the robust guidance provided by frame-wise conditions that are beneficial for interpolation involving large motion in linearts and animations \cite{zhu2024thin, Liu_2024_CVPR}.

\begin{table}[ht]\small
    \setlength{\abovecaptionskip}{3pt} 
  \setlength{\belowcaptionskip}{0pt}
  \setlength{\tabcolsep}{1.5mm}
  \caption{Ablation study on condition components.
  }
  \label{tab:ablation control}
  \centering
  \begin{tabular}{@{}lccccc@{}}
    \toprule
      & LPIPS ($\downarrow$) & FID ($\downarrow$)  & FVMD ($\downarrow$) & FVD ($\downarrow$)\\
    \midrule
    w/o Control & 0.2485 & 27.55 & 7217.5 & 536.5\\
    w/o Pose & 0.1843 & 24.70 & \textbf{5520.9} & 446.1 \\
    w/o Matching & 0.2124 & 24.17 & 6546.8 & 498.8 \\
    Full Model & \textbf{0.1832}& \textbf{24.05} & 5607.2 & \textbf{437.9}\\
  \bottomrule
  \end{tabular}
\end{table}

\subsection{Ablation Study}
We conduct ablations to to discuss the components of conditions and control weight $\gamma$. 

\begin{figure}[htb]
  \centering
    \setlength{\abovecaptionskip}{0pt} 
  \setlength{\belowcaptionskip}{0pt}
  \includegraphics[width=0.9\linewidth]{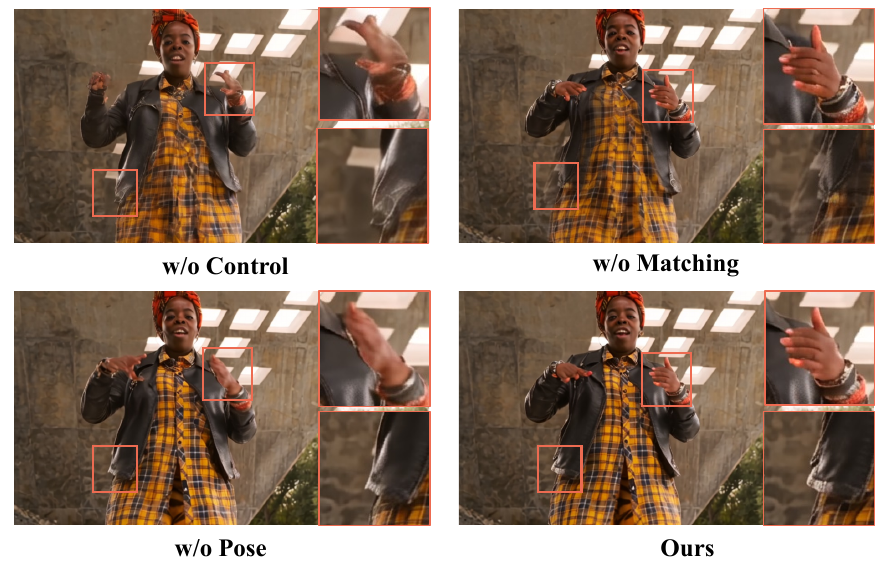}
  \caption{Ablation study on condition components. }
  \label{fig:ablation control}
  \vspace{-3mm}
\end{figure}

\noindent\textbf{Condition components.}
In \cref{tab:ablation control}, `w/o Control' denotes the exclusion of the entire frame-wise conditions control, while `w/o Pose' and `w/o Matching' indicate the removal of human pose and line matching conditions, respectively. The visual results are presented in \cref{fig:ablation control}, from which one can see that the line matching condition governs the overall motion of the scene, and the pose condition benefits details with human movements. 

\noindent\textbf{Control weight $\gamma$.}
The impact of $\gamma$ has been discussed in \cref{sec:flexibility}. Based on the findings in Figure \ref{fig:weight_compare} and Table \ref{tab:ablation weight}, FCVG is not very sensitive to the value of $\gamma$, and the weight $\gamma=1$ proves to be suitable for the majority of scenarios.

\begin{table}[htb]\small
    \setlength{\abovecaptionskip}{0pt} 
  \setlength{\belowcaptionskip}{0pt}
  \setlength{\tabcolsep}{1.5mm}
  \caption{{Quantitative analysis on control weight $\gamma$.}
  }
  \label{tab:ablation weight}
  \centering
  \begin{tabular}{@{}lccccc@{}}
    \toprule
      & LPIPS ($\downarrow$) & FID ($\downarrow$)  & FVMD ($\downarrow$) & FVD ($\downarrow$)\\
    \midrule
    $\gamma=0.5$ & 0.1912 & \textbf{23.80} & 5920.0 & \textbf{431.4} \\
    $\gamma=1.0$ & \textbf{0.1832}& 24.05 & \textbf{5607.2} & 437.9\\
    $\gamma=2.0$ & 0.1861 & 24.66 & 5726.9 & 441.1\\
  \bottomrule
  \end{tabular}
  \vspace{-3mm}
\end{table}

\section{Limitations and Future Work}

\begin{figure}[tb]
  \centering
    \setlength{\abovecaptionskip}{0pt} 
  \setlength{\belowcaptionskip}{0pt}
  \includegraphics[width=\linewidth]{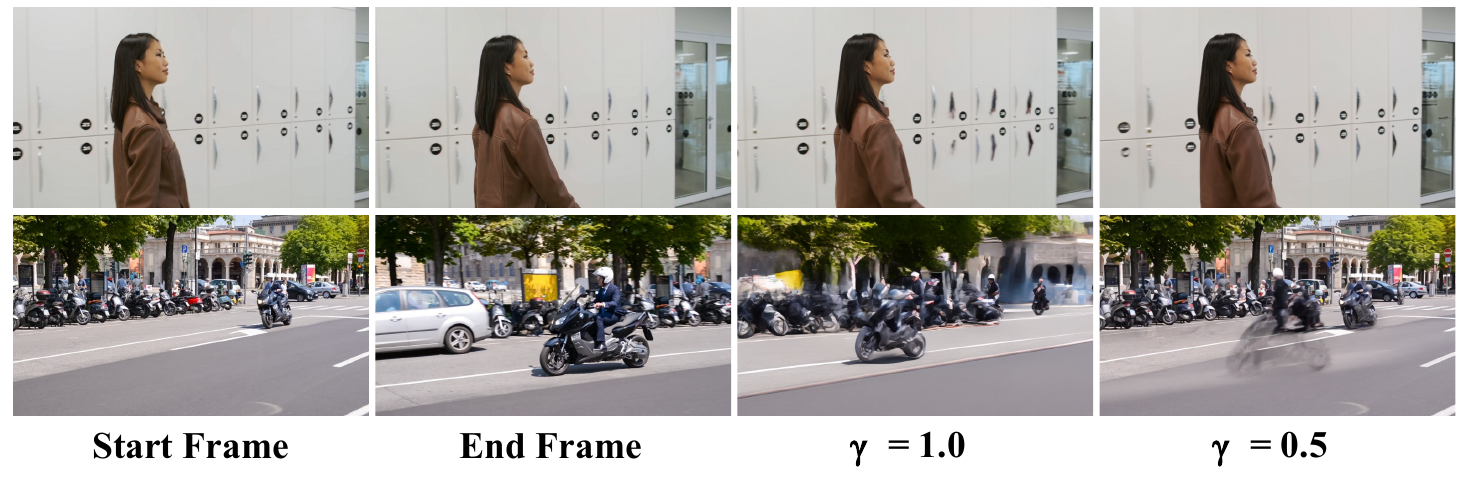}
  \caption{Failure cases. Incorrect matches and significant difference between input frames lead to intermediate artifacts, some of which can be mitigated by adjusting the control weight.}
  \label{fig:limitation}
  \vspace{-3mm}
\end{figure}

Due to the dependence on line matching, incorrect matching results may arise when two input frames exhibit highly similar features, thereby impacting the quality of generated frames. This limitation can be alleviated by manually reducing the control weight, as exemplified in the first instance in \cref{fig:limitation}. However, when there is a significant difference between the two input frames, matched lines may be sparse, making simple adjustment of the control weight ineffective, as the second example in \cref{fig:limitation}. This limitation could potentially be addressed in future research by replacing SVD with more robust I2V models. Furthermore, enhancing the interpolation process through diverse control conditions could offer improvements, e.g., incorporating user-specified drag effects like Dragdiffusion \cite{shi2024dragdiffusion} or generating control conditions based on generative models using user-defined text inputs \cite{wang2024holistic}.

\section{Conclusion}
In this paper, we propose a frame-wise conditions-driven video generation method, FCVG, for generative inbetweening. To fully exploit the potential of a video generation model in producing temporally stable videos, an explicit condition is provided for each frame. Specifically, frame-wise conditions can be obtained by interpolating two initial conditions that contain matched lines extracted from two input frames, and subsequently injecting them into the video generation model alleviates the ambiguity of inbetweening path. Our FCVG is not very sensitive to the introduced control weight, and thus exhibits robustness across diverse scenes with a fixed weight setting.
Moreover, FCVG is able to handle specific interpolation paths defined by users.

{
    \small
    \bibliographystyle{ieeenat_fullname}
    \bibliography{main}
}


\end{document}